%% file: bare_conf.tex
\documentclass[conference]{IEEEtran}
\ifCLASSINFOpdf
\else
\fi

\usepackage{times}
\usepackage{epsfig}
\usepackage{graphicx}
\usepackage{amsmath}
\usepackage{amssymb}
\usepackage[table]{xcolor}
\usepackage{color}
\usepackage{colortbl}
\usepackage{caption}
\usepackage{subcaption}
\usepackage{tabularx}
\usepackage{array}
\usepackage{multirow}
\usepackage{float}
\usepackage{amssymb} 
\usepackage{breakurl}

\begin{document}
%
\title{Deep Semantic Segmentation for Automated Driving: Taxonomy, Roadmap and Challenges}


\author{
\IEEEauthorblockN{Mennatullah Siam, Sara Elkerdawy, Martin Jagersand}
\IEEEauthorblockA{University of Alberta, Canada\\
Email: mennatul@ualberta.ca}

\and

\IEEEauthorblockN{Senthil Yogamani}
\IEEEauthorblockA{Valeo Vision Systems, Ireland\\
Email: senthil.yogamani@valeo.com}

}


%


\maketitle

%
\IEEEpeerreviewmaketitle

\begin{abstract}
\input{abstract}
\end{abstract}

\section{Introduction}
\label{sec:introduction}
\input{intro}

\section{Deep Semantic Segmentation Taxonomy} \label{sec:tax}
\input{survey}

\section{Deep Semantic Segmentation in Automated Driving} \label{sec:drive}
\input{problem}

\section{Challenges}\label{sec:chall}
\input{challenges}

\section{Alternative Application architectures }\label{sec:alternative}
\input{comparison}

\section{Benchmarking and Design Exploration}\label{sec:results}
\input{results}

\section{Conclusion}\label{sec:conc}
\input{conclusion}

\bibliographystyle{IEEEtran}
\bibliography{IEEEfull}


\end{document}

%% file: abstract.tex
Semantic segmentation was seen as a challenging computer vision problem few years ago. Due to recent advancements in deep learning, relatively accurate solutions are now possible for its use in automated driving. In this paper, the semantic segmentation problem is explored from the perspective of automated driving. Most of the current semantic segmentation algorithms are designed for generic images and do not incorporate prior structure and end goal for automated driving. First, the paper begins with a generic taxonomic survey of semantic segmentation algorithms and then discusses how it fits in the context of automated driving. Second, the particular challenges of deploying it into a safety system which needs high level of accuracy and robustness are listed. Third, different alternatives instead of using an independent semantic segmentation module are explored. Finally, an empirical evaluation of various semantic segmentation architectures was performed on CamVid dataset in terms of accuracy and speed. This paper is a preliminary shorter version of a more detailed survey which is work in progress.

%% file: intro.tex
Semantic image segmentation has witnessed tremendous progress recently with deep learning. Semantic segmentation is targeted towards partitioning the image into semantically meaningful parts with various applications for that. It has been used in robotics \cite{valada2016deep}\cite{bonanni2013human}\cite{vineet2015icra}\cite{kundu2014joint}, medical applications \cite{cciccek20163d}\cite{zhu2016adversarial}, augmented reality \cite{miksik2015semantic}, and most prominently automated driving \cite{zhang2013understanding}\cite{ros2016synthia}\cite{brostow2009semantic}\cite{cordts2016cityscapes}. 

Automated driving is another hot topic in the literature that is advancing with the rapid growth in deep learning. The goal to create an automated car started since 1989 with the work in \cite{pomerleau1989alvinn} that used single hidden layer network. However, the limitations in neural networks at that time did not allow its progress further on. Recently with deep learning and advances in GPU technologies, different works on automated driving emerged. 

Two main paradigms for automated driving emerged: \textbf{(1)} The mediated perception approach which parses the whole scene and uses this information for the control decision increasing the complexity and the cost of the system. \textbf{(2)} The behavior reflex paradigm that relies more on end-to-end learning to map direct sensory input to driving decision which is an ill-posed problem due to the many possible ambiguous decisions, such as the work in \cite{bojarski2016end}\cite{muller2005off}. However, in \cite{chen2015deepdriving} an intermediate approach was suggested that learns affordance indicators for the driving scene. These indicators can then feedback on a simple controller for the final driving decision. The previous work on automated driving pose the important question of whether the solution for automated driving need semantic segmentation module or not? 

A related survey in \cite{zhu2016beyond} on semantic segmentation literature is presented. However it is not addressing the specific application of automated driving. This paper tries to address this gap by reviewing the work on semantic segmentation in the context of automated driving. This paper addresses the question on what is the importance of semantic segmentation in automated driving and reviews alternative approaches. The paper is organized as follows, section \ref{sec:tax} covers the literature work on deep semantic segmentation in general. Followed by section \ref{sec:drive} that focuses on the problem of automated driving and how can semantic segmentation be used in it. Then section \ref{sec:chall} presents the main challenges in automated driving applications. Then alternative approaches are discussed including end-to-end learning and multi-task learning in section \ref{sec:alternative}. A comparative evaluation of semantic segmentation architectures is presented in \ref{sec:results}. Finally section\ref{sec:conc} summarizes and presents the conclusions.

%% file: survey.tex
In this section the categories of deep semantic segmentation are discussed. Different work under these categories are reviewed in further details with discussion of their limitations if any and future directions. The literature work in semantic segmentation is categorized into four subcategories: \textbf{(1)} Classical Methods. \textbf{(2)} Fully Convolutional Networks. \textbf{(3)} Structured Models. \textbf{(4)} Spatio-Temporal Models. The first category reviews the classical approaches before the emergence of deep learning. The second category is about the main body of work on semantic segmentation using deep learning. The third category reviews the work that tries to utilize structure in the problem of semantic segmentation. Thus following the assumption that neighboring pixel labels should be coherent. Then the fourth category exploits the temporal information that is present in videos. Table \ref{table:taxonomy} shows the detailed taxonomy of semantic segmentation approaches.

\begin{table*}[ht!]
\centering
\caption{Taxonomy of Semantic Segmentation Architectures}
\label{table:taxonomy}
\begin{tabular}{|l|l|l|l|}
\hline
\textbf{Classical Methods} & \textbf{Fully Convolutional Networks}                                                 & \textbf{Structured Models} & \textbf{Spatio-Temporal Models} \\ \hline
Decision Forests \cite{shotton2008semantic}\cite{brostow2008segmentation} & Patchwise Training \cite{farabet2013learning}\cite{farabet2012scene}\cite{grangier2009deep} & Merged with CRFs \cite{lin2015efficient}\cite{chen2016deeplab}\cite{zheng2015conditional} & Clockworks Net \cite{ShelhamerRHD16}\\ \hline

CRFs \cite{sturgess2009combining}\cite{russell2009associative} & Pixelwise Training \cite{long2015fully}\cite{noh2015learning}\cite{badrinarayanan2015segnet} & Using RNNs \cite{visin2016reseg} &  Using RNNs \cite{FayyazSSFK16}\cite{siam2016convolutional}\cite{nilsson2016semantic}\\ \hline

Boosting \cite{sturgess2009combining}\cite{shotton2006textonboost} & Multiscale \cite{yu2015multi}\cite{farabet2013learning}\cite{noh2015learning}\cite{chen2015attention}\cite{Qi_2016_CVPR}\cite{ronneberger2015u} &  & \\ \hline
\end{tabular}
\end{table*}

\subsection{Classical methods}
Few years ago, semantic segmentation was seen as a challenging problem to achieve reasonable accuracy. The main approaches used in semantic segmentation was based on random forest classifier or conditional random fields. In \cite{shotton2008semantic} decision forests were used, where each tree was trained on random subset of the training data. These methods implicitly cluster the pixels while explicitly classifying the patch category. In \cite{brostow2008segmentation} a randomized decision forest was also used however instead of using appearance based features, motion and structure features were used. These features include surface orientation, height above camera, and track density where faster moving objects have sparser tracks than static objects. However, these techniques rely on hand crafted features and perform pixel-wise classification independently without utilizing the structure in the data. 

On the other hand conditional random fields(CRF) were proven to be a good approach for structured prediction problems. In \cite{sturgess2009combining}\cite{russell2009associative} segmentation is formulated as CRF problem. The energy function used in CRF formulation usually contains unary potential and pairwise potential. The unary potential gives a probability of whether the pixel belongs to a certain class. While pairwise potential which is also referred to as smoothness term ensures label consistency among connected pixels. Boosting is another method that can be used to classify pixels. It is based on combining multiple weak classifiers that are based on some shape filter responses, as in \cite{sturgess2009combining}\cite{shotton2006textonboost}. However the progress in classical methods was always bounded by the performance of the hand crafted features used. But that was overcome with deep learning as will be discussed in the following sections.

\subsection{Fully Convolutional Networks(FCN)}
The area of semantic segmentation using convolutional neural networks witnessed tremendous progress recently. There were mainly three subcategories of the work that was developed. The first \cite{farabet2013learning}\cite{farabet2012scene}\cite{grangier2009deep} used patch-wise training to yield the final classification. In\cite{farabet2013learning} an image is fed into a Laplacian pyramid, each scale is forwarded through a 3-stage network to extract hierarchical features and patch-wise classification is used. The output is post processed with a graph based classical segmentation method. In \cite{grangier2009deep} a deep network was used for the final pixel-wise classification to alleviate any post processing needed. However, it still utilized patch-wise training. 

\begin{figure}[!t]
\centering
\includegraphics[width=0.5\textwidth]{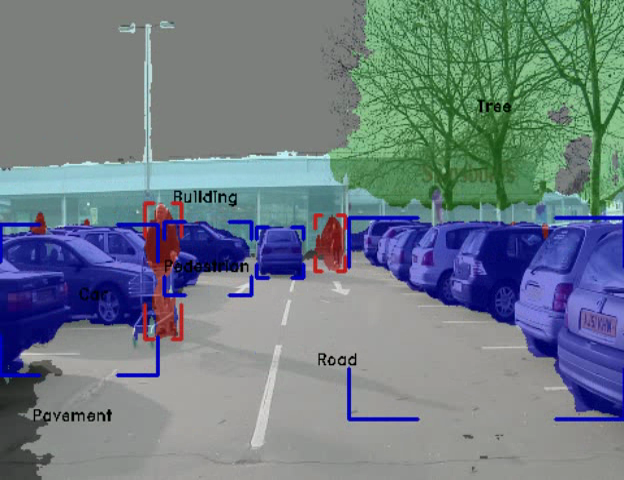}
\caption{Semantic Segmentation of a typical automotive scene}
\label{segnet}
\end{figure}

The second subcategory \cite{long2015fully}\cite{noh2015learning}\cite{badrinarayanan2015segnet} was focused on end-to-end learning of pixel-wise classification.  It started with the work in \cite{long2015fully} that developed fully convolutional networks(FCN). The network learned heatmaps that was then upsampled with-in the network using deconvolution to get dense predictions. Unlike patch-wise training methods this method uses the full image to infer dense predictions.  In \cite{noh2015learning} a deeper deconvolution network was developed, in which stacked deconvolution and unpooling layers are used. In Segnet \cite{badrinarayanan2015segnet} a similar approach was used where an encoder-decoder architecture was deployed. The decoder network upsampled the feature maps by keeping the maxpooling indices from the corresponding encoder layer. In Figure \ref{segnet} an example of the semantic segmentation output of segnet applied in an automated driving setting is shown.

Finally, the work in \cite{yu2015multi}\cite{farabet2013learning}\cite{noh2015learning}\cite{chen2015attention}\cite{Qi_2016_CVPR}\cite{ronneberger2015u} focused on multiscale semantic segmentation. Initially in \cite{farabet2013learning} the scale issue was addressed by introducing multiple rescaled versions of the image to the network. However with the emergence of end-to-end learning, the skip-net architecture in \cite{long2015fully} was used to merge heatmaps from different resolutions. Since these architectures rely on downsampling the image, loss of resolution can hurt the final prediction. The work in \cite{ronneberger2015u} proposed a u-shaped architecture network where feature maps from different initial layers are upsampled and concatenated for the next layers. Another work in \cite{yu2015multi} introduced dilated convolutions, which expanded the receptive field without losing resolution based on the dilation factor. Thus it provided a better solution for handling multiple scales. Finally the recent work in \cite{chen2015attention} provided a better way for handling scale. It uses attention models that provides a mean to focus on the most relevant features with-in the image. This attention model is able to learn a weighting map that weighs feature maps pixel-by-pixel from different scales.

\subsection{Structured Models}
The previous approaches in fully convolutional networks do not utilize the structure in the data. Thus, recent work was directed towards using the prior structure in the data. Specifically in automotive scenes prior structure can be exploited for better segmentation. The commonly used model to incorporate structure is conditional random field (CRF) \cite{lin2015efficient}\cite{chen2016deeplab}\cite{zheng2015conditional}. In \cite{lin2015efficient}, CRF is used as a post processing step after the segmentation network. In \cite{chen2016deeplab}, CRF is also used as post processing to a dilated convolution network to take contextual information into consideration. Finally, in \cite{zheng2015conditional} the mean field inference algorithm that is used within CRF formulation was formulated as a recurrent network.

Another way to model structure is by using a recurrent neural network (RNN) to capture the long range dependencies of various regions \cite{visin2016reseg}. It introduced a different formulation for solving the structured prediction problem. A Recurrent layer is used to sweep the image horizontally and vertically, which ensures the usage of contextual information for a better segmentation.

\subsection{Spatio-Temporal Models}
All the discussed work was focused on still image segmentation. Recently some approaches emerged for video semantic segmentation that utilized temporal information \cite{ShelhamerRHD16}\cite{FayyazSSFK16} \cite{siam2016convolutional}\cite{nilsson2016semantic}. In \cite{ShelhamerRHD16} introduced clockworks which are clock signals that control the learning of different layers with different rates. In \cite{FayyazSSFK16} spatio temporal FCN is introduced by using a layer grid of Long Short term memory models (LSTMs). However conventional LSTMs do not utilize the spatial coherence and would end up with more parameters to learn. 

In a recent work \cite{siam2016convolutional} convolutional gated recurrent networks was used to learn temporal information to leverage the semantic segmentation of videos. The recurrent unit used in this work was convolutional which enables it to learn both spatial and temporal information with less number of parameters. Thus, it was easier to train and memory efficient. The work in \cite{nilsson2016semantic} combined the power of both convolutional gated architectures and spatial transformers for leveraging video semantic segmentation.

%% file: problem.tex
\subsection{Problem Structure}
Semantic segmentation for automated driving has many a priori constraints relative to a general version. In this section, we discuss the various aspects which brings a simplifying structure to the problem.

\subsubsection{Scene Structure}
Prior information could simplify model complexity greatly. There are different types of prior information that can be used. Spatial priors such as the fact that lanes lie on a ground plane, or that road segmented is mostly in the bottom half of the images. Geometric priors on the shapes of objects, for examples lanes are thick lines that are all converging into a vanishing point. Color priors such as the color of traffic lights or white lanes. Finally, Location priors, for example the lane, road or buildings locations based on high definition maps or aerial maps.

\subsubsection{Multi-camera Structure}
\begin{figure}[!t]
\centering
\includegraphics[width=0.5\textwidth]{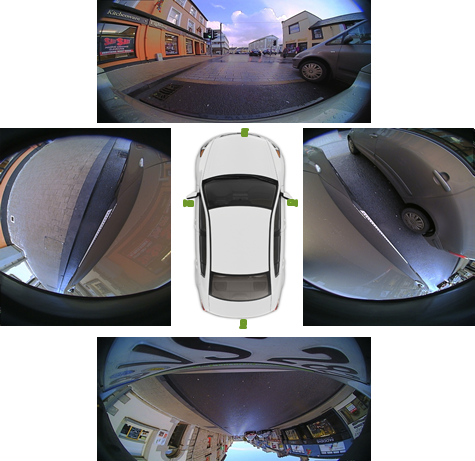}
\caption{Sample images from a surround-view camera network }
\label{fig:surround}
\end{figure}

Typically automotive systems uses a multi-camera network. Current systems have at least four cameras and it is increasing to more than ten cameras for future generation systems. Figure \ref{fig:surround} shows sample images of the four cameras mounted on the car. It covers the entire 360$^{\circ}$ field of view surrounding the car. The geometric structure of the four cameras and the motion of the car induces a spatio-temporal structure across the four images. For example, when the car is turning left, the region imaged by the front camera will be imaged by the right-mirror camera after a delay. There is also similarity in the near-field road surface in all the four cameras as they belong to the same road surface.

\subsection{Dense High Definition(HD) maps}

\begin{figure}[!t]
\centering
\includegraphics[width=0.5\textwidth]{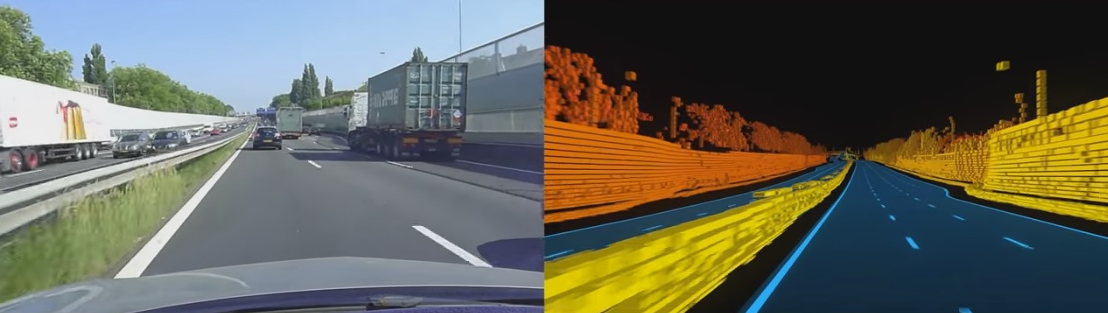}
\caption{Example of High Definition (HD) map from TomTom RoadDNA (Reproduced with permission of the copyright owner) }
\label{fig:HDmaps}
\end{figure}

High accuracy of Object detection is very difficult to achieve and HD maps is an important cue to improve it. There are two types of HD maps: \textbf{(1)} Dense Semantic Point Cloud Maps and \textbf{(2)} Landmark based Maps. The former is the dense version where the entire scene is modeled by 3D point cloud with semantics. Google and TomTom adopt this strategy. As this is high end, it is expensive to cover the entire world and needs large memory requirements.  If there is good alignment, all the static objects (road, lanes, curb, traffic signs) are obtained from the map already and dynamic objects are obtained through background subtraction. TomTom RoadDNA\cite{TomTomRoadDNA} provides an interface to align various sensors like Lidar, Cameras, and others. Figure \ref{fig:HDmaps} illustrates this where the pre-mapped semantic point cloud on the right is aligned with an image at run-time with other dynamic objects. They have mapped majority of European cities and their system provides an average localization error within 10 cm assuming a coarse location from GPS. Landmark based maps are based on semantic objects instead of generic 3D point clouds. Thus it works primarily for camera data. Mobileye and HERE follow this strategy. This can be viewed as a simple form of the 3D point cloud where a subset of objects is mapped using a 2D map. In this method, object detection is leveraged to provide a HD map and the accuracy is improved by aggregating over several observations from different cars. 

In case of a good localization, HD maps can be treated as a dominant cue and semantic segmentation algorithm greatly simplifies to be a refinement algorithm of priors obtained by HD maps. In Figure \ref{fig:HDmaps}, the semantic point cloud alignment provides an accurate semantic segmentation for static objects. Note that it does not cover distant objects like sky. This would need a good confidence measure for localization accuracy, typically some kind of re-projection error is used. HD maps can also be used for validation or post-processing the semantic segmentation to eliminate false positives. For this, both landmark maps and semantic point cloud maps could be used. 

\subsection{Localization}
Localization or depth estimation is very critical for automated driving. Having image semantics without localization is not very useful. 

\subsubsection{Depth using Structure from Motion(SFM)}
The straight forward approach to augment localization is to have a parallel independent path for computing dense depth using a standard method like structure from motion (SFM) and then augmenting the depth to localize the objects. Dense depth is computed to understand the spatial geometry of the scene. Accurate Depth should help in semantic segmentation and could be passed on as an extra channel. However, SFM estimates are quite noisy and also the algorithm variations over time could affect the training of the network. But in \cite{brostow2008segmentation} some cues from the noisy point-cloud was inferred to act as features for segmentation. The cues proposed were: height above the  camera,  distance  to  the  camera  path,  projected surface orientation, feature track density, and residual reconstruction error. The work in\cite{kundu2014joint} proposed a way of jointly estimating the semantic segmentation and structure from motion in a conditional random field formulation.

\subsubsection{LIDAR sensors}

LIDAR sensors provide very accurate depth estimation. However, they are not dense in the image lattice. This leads to problems in learning a dense convolutional neural networks features. But it can provide a way to fuse semantic segmentation with depth information in a probabilistic framework. In \cite{mccormac2016semanticfusion} the method fused a map built using elastic fusion \cite{whelan2015elasticfusion} and semantic segmentation from convolutional neural networks termed as semantic fusion. The class probabilities were maintained for each pixel in the map and updated in an incrementally Bayesian method. The images used in this work were from RGB-D cameras, but it provided potential use of depth from LIDAR sensors. Generally, this is a good research problem to be pursued as LIDAR is becoming a standard sensor for next generation automated driving systems. 


\subsubsection{Joint In-the-Network Localization}
There exists promising algorithms on using convolutional neural networks to estimate structure and camera motion. A recent work in \cite{ummenhofer2016demon} proposed depth and motion network for learning monocular stereo. As far as the authors are aware, there is no work on jointly estimating depth and semantics with in a network. This can synergize and potentially aid in the estimation of each other. It can also be trained simultaneously in an end-to-end fashion. This problem can be of potential future direction for further research.

\begin{figure}[!t]
\centering
\includegraphics[width=0.5\textwidth]{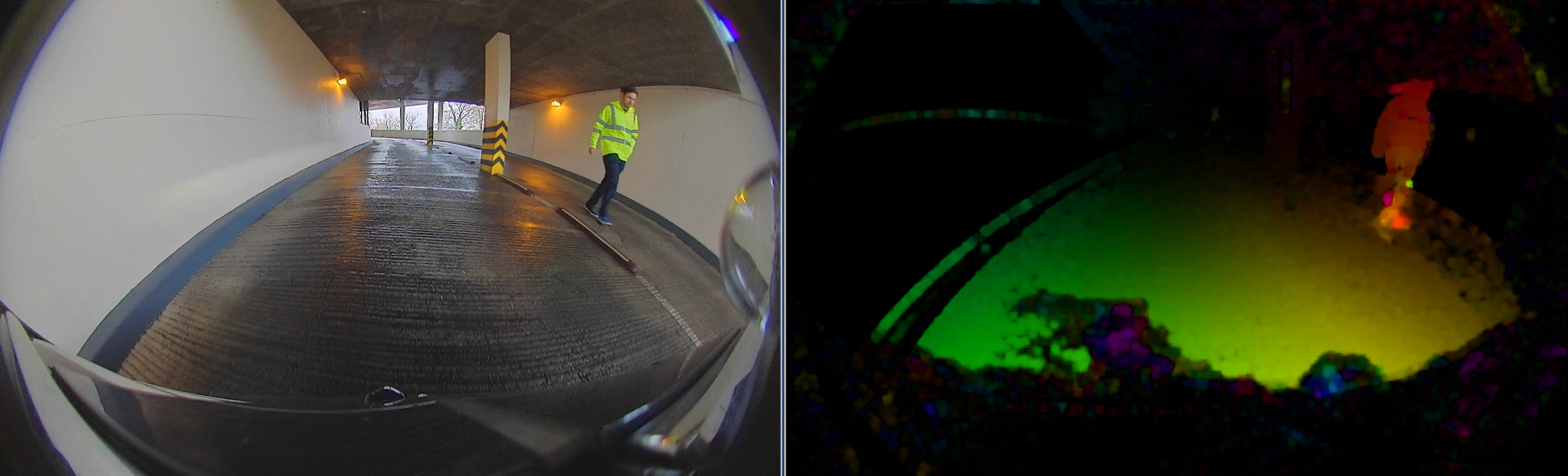}
\caption{Illustration of dense optical flow from which dense depth for SFM can be obtained}
\label{fig:denseSFM}
\end{figure}

%% file: challenges.tex
\subsection{Computational Bound in Embedded Systems}
On a high end automotive platform like Nvidia Tegra X1, Enet \cite{paszke2016enet} achieves around 4 fps and the proposed algorithm in \cite{tremlspeeding} achieves around 3 fps at a slightly higher accuracy. This benchmark is for a 720P resolution and the current generation cameras are around 2 Megapixel which will reduce the runtime by another factor of 3X. This is clearly not acceptable for a commercial solution to handle high speed objects for highway driving. Reducing the resolution to VGA (640x480) brings it close to 10 fps which is still not reasonable and reducing resolution degrades accuracy and misses small objects which might be critical. Additionally, for full surround view sensing at least 4 cameras need to be employed which adds in another factor of 4X. However the industry is moving towards custom hardware accelerators for CNNs which will enable the possibility of doing multi-camera semantic segmentation at a higher frame rate, Nvidia Xavier for instance supports 30 tera-ops. There is also active research on efficient network design which will improve the performance.

\subsection{Need of large annotated datasets}
The real potential of deep learning was unveiled because of the large dataset Imagenet\cite{imagenet_cvpr09}. The functional complexity of semantic segmentation is much higher and it requires a significantly larger dataset relative to Imagenet. Annotation for semantic segmentation is time consuming, typically it can take around an hour for annotating a single image. It can be speeded up by the availability of other cues like LIDAR or exploiting temporal propagation and bootstrapping classifier.

The popular semantic segmentation automotive datasets are CamVid \cite{brostow2008segmentation} and the more recent cityscapes \cite{cordts2016cityscapes}. The latter has a size of 5000 annotation frames which is relatively small. The algorithms trained on this dataset do not generalize well to data tested on other cities and with unseen objects like tunnels. To compensate for that, synthetic datasets like Synthia \cite{ros2016synthia} and Virtual KITTI \cite{gaidon2016virtual} were created. There is some literature which demonstrates that a combination produces reasonable results in small datasets. But they are still limited for a commercial deployment of an automated driving system. 

Hence there is a recent effort to build larger semantic segmentation datasets like Mapillary Vistas dataset \cite{mapillary} and Toronto City \cite{wang2016torontocity}. Mapillary dataset comprises of 25,000 images with 100 classes. It also offers large variability in terms of weather condition, camera type and geographic coverage. Toronto City is a massive semantic segmentation, mapping and 3D reconstruction dataset covering 712 km$^{2}$ of land, 8439 km of road and around 400,000 buildings. The annotation is completely automated by leveraging Aerial Drone data, HD maps, city maps and LIDARs. It is then manually verified and refined. 

\subsection{Learning Challenges}
\subsubsection{\textbf{Class imbalance}} There is severe class imbalance due to the fact that important objects like pedestrians are under represented unlike sky and building. This could also create a bias to ignore small objects. This could be handled by a weighting scheme in the error function. Another potential solution is to use mask predictions on detected bounding boxes of these small objects as in \cite{DBLP:journals/corr/DaiHLRS16}\cite{teichmann2016multinet}. 

\subsubsection{\textbf{Unobserved Objects}} Because the soft-max classifier is normalized to probability one, it doesn't handle previous unseen objects. The classifier matches it to one of the previously trained classes. It is not possible to cover all possible objects in training phase (eg: a rare animal like Kangaroo or a rare vehicles like construction truck). This could be handled by measuring uncertainty of the output classification, similar to Bayesian Segnet \cite{DBLP:journals/corr/KendallBC15}.  

\subsubsection{\textbf{Complexity of Output}} The output representation of semantic segmentation is a set of complex contours and can be very complex in very high textured scenes. The post processing modules like mapping or maneuvering require a much simpler representation of objects. This leads to a question of learning to classify this simpler representation directly instead of semantic segmentation.

\subsubsection{\textbf{Recovering individual objects}} Pixel-wise Semantic segmentation produces regions of same object and hence does not provide individual objects in a segment. This might be needed for tracking applications which tend to track objects like pedestrians individually. One solution is to use post processing classifier to further sub-divide the regions but this could be directly classified instead. However, a recent instance level segmentation paradigm can segment different instances of the same class as in \cite{DBLP:journals/corr/DaiHLRS16} without the need for post processing.

\subsubsection{\textbf{Goal Orientation}} Semantic segmentation is a generic problem and at the moment there is no goal orientation towards the end goal of automated driving. For example, there may not be a need for accurate contour of objects or in detecting irrelevant objects like sky for end driving goal. This could be achieved by customizing the loss function (eg: weighting of important objects) but a modular end to end system will be scalable to automatically perform it. 

\subsubsection{\textbf{Variable object complexity}} A typical automotive scene has large  complexity variability with simple structures like road or sky and complex structures like pedestrians. Pedestrians have higher complexity due to large appearance variations and articulations. Thus instead of using a small complexity network across the image, a variable complexity network like a cascaded CNN \cite{li2017not} will be more efficient. 

\subsubsection{\textbf{Corner Case Mining}} As the object detection parts are tightly coupled, it is difficult to do hard negative mining and to analyze corner cases. Even when the corner cases are known conceptually, it can be hard to record video sequences for the same. Synthetic sequences could be used to design such scenarios.

%% file: comparison.tex
In this section different alternatives to pure semantic segmentation are discussed. We present it with other possibilities where it can be coupled.

\textbf{Multi-task Learning:} Since the same CNN features generalize well for various tasks beyond object detection like flow estimation, depth, correspondence, and tracking. Thus a common CNN feature pipeline can be harmonized to be used for various tasks. In \cite{teichmann2016multinet}, they propose a CNN encoder and decoder for various tasks like scene classification and vehicle detection. A joint flow estimation and semantic segmentation in \cite{DBLP:journals/corr/HurR16} was presented. 

\begin{figure}[!t]
\centering
\includegraphics[width=0.5\textwidth]{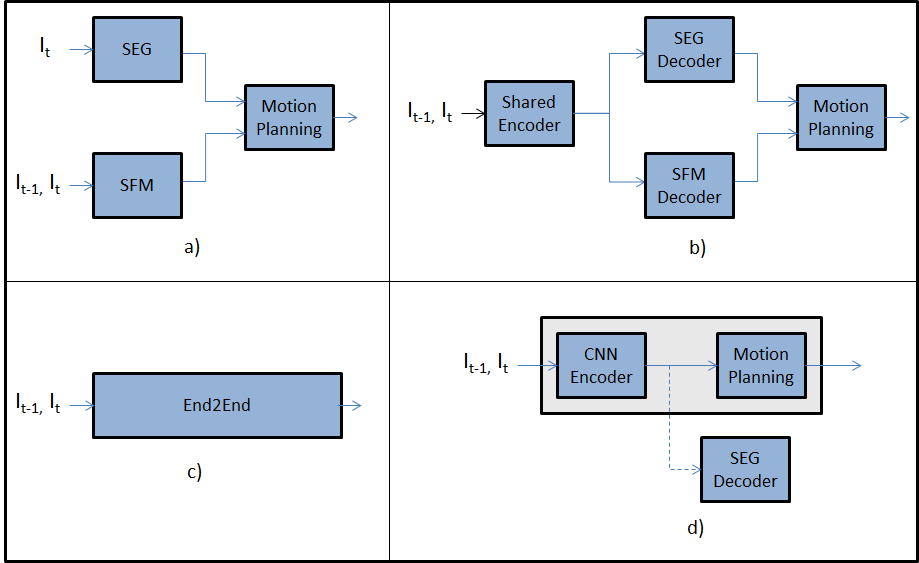}
\caption{Different application architectures - a) Classical architecture b) Shared encoder for multi-task learning c) End to end learning d) Modular end to end learning}
\label{multinet}
\end{figure}

\textbf{End to end learning:} Human beings perform soft computing and they do not perform an accurate object detection when driving. They are goal oriented and an accurate object detection is not necessary for safe driving. End to end has the big advantage of not having to do any annotation as the driving signal outputs are obtained directly from the Controller Area Network (CAN) signals. Companies like Uber are betting this away as they can collect lots of driving data through their taxi fleet. 

The output is of fewer dimensions (brake, steering, acceleration) and also temporally smooth. Hence for the same input, mathematically this function should have a simpler functional complexity relative to the complex output structure of semantic segmentation. The work in \cite{bojarski2016end}\cite{muller2005off} is in that direction.

\textbf{Modular End to End learning:} We use the term modular end to end learning when there are auxiliary losses to ensure safety and interpret ability. For instance, segmentation loss can be added as an auxiliary loss for an end to end driving CNN \cite{xu2016end}. Using this auxiliary loss, the CNN loosely learns to semantically segment, but it is also learns to have a better representation for the intermediate features. It was shown in that work that using auxiliary loss outperforms the vanilla end to end learning. The work also uses recurrent gated unit after the CNN to model temporal information.



%% file: results.tex
\subsection{Benchmarking}
In this section a comparative evaluation of different semantic segmentation architectures is presented. Although there has been numerous work showing evaluation of different architectures on CamVid \cite{brostow2008segmentation}. However the previous work was concerned only with the accuracy of the segmentation. We present an evaluation of both shallower and state of the art work in terms of mean intersection over union and speed. The comparison is shown in Table \ref{table:camvid}. Some networks that has not been evaluated on the semantic segmentation for automated driving are also presented. Thus covering a wider range of potential efficient architectures. This can guide further decisions on what would best fit in the automated driving system. Although other architectures such as DeepLab\cite{chen2016deeplab} show much better accuracy and are the state of the art in segmentation, but are computationally inefficient. Thus, these architectures are not included in the comparison. Evaluation metrics used are mean intersection over union(IoU) and per class IoU. The running time for inference is computed in seconds. The different architectures are evaluated on a GTX TITAN GPU with images of resolution 480x360. 

The architectures that are primarily evaluated are : (1) Unet \cite{ronneberger2015u}. (2) Xception \cite{chollet2016xception} which is a classification network that was not used in the segmentation problem before. (3) Dilated FCN16s, an architecture that was designed to be computationally and memory efficient with reasonable accuracy. (4) FCN8s \cite{long2015fully}. (5) Segnet Basic \cite{badrinarayanan2015segnet}. (6) Dilation8 \cite{yu2015multi}. (7)Enet \cite{paszke2016enet}, which is the most efficient architecture for semantic segmentation. A unified framework with the first five architectures is going to be publicly available to help further research. While the results of the last two architectures are reported from their work. Note that the mean class IoU is computed over all classes even the ones not included in the Table \ref{table:camvid}. But, only the classes of interest were the ones mentioned in Table \ref{table:camvid}.

Although Dilation8 outperforms all previous architectures in mean IoU it has the largest running time. This renders it as an inefficient solution to semantic segmentation for automated driving. However, the dilated convolution idea can be adapted in a shallower network. It uses dilated convolution to increase receptive field while maintaining the resolution of the segmentation. Dilated FCN16s is an adapted version of FCN-16s as originally introduced in \cite{long2015fully}. Two pooling layers are removed along with the convolutional layers in between them and conv4/conv5 layers are reduced to two dilated convolution layers with dilation factors of 2 and 4 respectively. This leads to a decrease in the size of the network and its running time for real-time applications. Another architecture used for medical image segmentation was experimented on CamVid which is called Unet. It turned to work second best on CamVid, but the running time is still not practical for real deployment. Xception \cite{chollet2016xception} is an architecture that is mainly relying on depthwise separable convolution, that separates the spatial convolution from depthwise convolution. Although the network is designed for classification, it has been transformed to a fully convolutional network for the purpose of segmentation. The network mean IoU was much lower than other architectures, with a very small improvement in the running time against Segnet. Although Segnet is not considered as the state of the art in segmentation, but it turned out to provide a good balance between mean IoU and speed. In our experiments using batch normalization \cite{ioffe2015batch} turned to be effective in training both Segnet and Unet. It turned to converge faster, and it got better mean IoU of 47.3\% in case of Segnet. It is worth noting that in case of FCN8s, we were able to reproduce similar results to the work in\cite{FayyazSSFK16}. But this is less by 2\% than what was reported in \cite{DBLP:journals/corr/KendallBC15}. 


\begin{table*}[ht!]
\centering
\caption{Semantic Segmentation Results on CamVid. Running time in seconds, mean IoU, and perclass IoU is shown. Some of the 11 classes are shown due to limited space. }
\label{table:camvid}
\begin{tabular}{|c|c|c|c|c|c|c|c|c|c|}
\hline
\multirow{2}{*}{ } & \multirow{2}{*}{Run-time (s)} & \multirow{2}{*}{Mean Class IoU} & \multicolumn{7}{c|}{Per-Class IoU}\\ \cline{4-10}  
& & &  Sky & Building & Road & Sidewalk & Vegetation & Car & Pedestrian \\ \hline
FCN16s Dilated & 0.07 & 46.7 & 86.3 & 69.1 & 87.8 & 63.7 & 60.8 & 63.6 & 21.4\\ \hline
Xception\cite{chollet2016xception} & \textbf{0.02} & 42.8 & 81.9  &  68.9 & 86.6 &  62.9 & 61.6 & 60.8 & 19.8\\ \hline
Segnet-Basic\cite{badrinarayanan2015segnet}  & 0.03 & 46.4 & 87.0 & 68.7 & 86.2 & 60.5 & 52.0 & 58.5 & 25.3 \\ \hline
FCN8s\cite{long2015fully} & 0.33  & 49.7 & 87.6 &  75.5 & 87.2 & 67.2 & 70.6 & 76.4 & 27.7 \\ \hline
Unet\cite{ronneberger2015u}+BN\cite{ioffe2015batch} & 0.56 & 53.9 & 90.2 & 72.6 & 89.1 & 67.2 & 67.7 & 74.7 & 34.1 \\ \hline
Dilation8\cite{yu2015multi} & 0.6474 & \textbf{65.3} & 89.9 &  \textbf{82.6} & 92.2 & 75.3 & 76.2 & \textbf{84.0} & 56.3 \\ \hline
Enet\cite{paszke2016enet} & 0.047 & 51.3 & \textbf{95.1} &  74.7 & \textbf{95.1} & \textbf{86.7} & \textbf{77.8} & 82.4 & \textbf{67.2} \\ \hline
\end{tabular}
\end{table*}

\subsection{Design Exploration}

Deep learning is a rapidly progressing area of research. Most of the research is disparate in which the various ideas developed in different architectures are not formalized because of the lack of theory. Hence it is hard to combine ideas from two top networks from an application development perspective. Additionally the main area of active research in deep learning is on image recognition problems (as in ImageNet challenge) and the ideas trickle down to semantic segmentation. Additionally efficiency is typically not a design criteria in academic research as majority of leading networks are very large comprising of hundreds of layers and employ ensemble of several networks. The work in \cite{canziani2016analysis} compared various networks' accuracy normalized to the amount of computation and shows ResNet and GoogleNet are efficient architectures. This suffers from the same problem of treating the different networks independently and hence does not formalize and combine ideas. 

Some good design choices that are accepted with-in the community are presented:\\
\textbf{(1)} The use of 3x3 convolutions similar to VGG architectures \cite{simonyan2014very} turned to be useful experimentally. Especially in scenarios where you care about the resolution of your input such as segmentation. Since larger filter size will cause reduction in the image resolution.\\
\textbf{(2)} The dilated convolution is considered to be the best practice in segmentation as it increases the receptive field without downgrading resolution. Although in our comparative evaluation Dilation8 was not suitable for real-time applications. However that can be due to their use of a deep network to build upon.\\
\textbf{(3)} For real-time performance shallow networks can be useful for segmentation with a compromise in the accuracy.\\
\textbf{(4)} Batch normalization \cite{ioffe2015batch} turned to be a very useful trick for better convergence during training in our experiments. This is due to the reduction of change in distribution of network activations. They termed that as reduction of internal covariate shift. The covariate shift occurs due to change of networks parameters during training.\\
\textbf{(5)} The resolution of the input image largely affects the segmentation, although it seems as a tiny detail. We found that higher input image resolution can help with segmenting small objects like pedestrian. Also, using random crops to help reduce the class imbalance can further help the segmentation. This can be seen in Dilation8 results, they use random crops from the image that is then upsampled as input.\\
\textbf{(6)} Skip connections is widely used in segmentation architectures such as FCN8s \cite{long2015fully} and U-net \cite{ronneberger2015u}. However, the extensive use of these skip connections can lead to overhead in memory bandwidth. \\


%% file: conclusion.tex
In this paper, we have conducted a detailed review of deep semantic segmentation from the perspective of automated driving. First, we provide a generic survey of various architectures including fully convolutional network architectures and different variants that work patch-wise, pixel-wise, or that support multi-scale aggregation. Then we discuss the use of structured models and spatio-temporal features in the segmentation problem is discussed. Second, we reviewed various semantic segmentation datasets that can be used for automated driving setting and presented a set of challenges to orient both the research and industrial community towards the current bottlenecks. Finally, a comparative evaluation is conducted on different state of the art network architectures on an urban scene dataset including both runtime and accuracy. The comparison included architectures that have not been tried for automated driving scenes. Due to page limitations, the contents of this survey are kept high level. The authors are working on a more detailed survey which will include more detailed  analysis and discussion.